\documentclass[letterpaper]{article} 
\usepackage{aaai2026}  
\usepackage{times}  
\usepackage{helvet}  
\usepackage{courier}  
\usepackage[hyphens]{url}  
\usepackage{graphicx} 
\usepackage{natbib}  
\usepackage{caption} 
\usepackage{algorithm}
\usepackage{algorithmic}

\usepackage{amsmath}
\usepackage{amssymb}
\usepackage{booktabs}
\usepackage{newfloat}
\usepackage{listings}
\usepackage{subcaption}
\usepackage{multirow}
\usepackage{natbib}

\newcommand{\eat}[1]{}
\newcommand{\eg}{{\em e.g.,~}}     
\newcommand{\ie}{{\em i.e.,~}}      

\usepackage{newfloat}
\usepackage{listings}
\DeclareCaptionStyle{ruled}{labelfont=normalfont,labelsep=colon,strut=off} 
\floatstyle{ruled}
\newfloat{listing}{tb}{lst}{}
\floatname{listing}{Listing}
\title{Graph Smoothing for Enhanced Local Geometry Learning\\in Point Cloud Analysis}
\author{
    Shangbo Yuan$^{1}$, Jie Xu$^{2,}$\footnote{Corresponding Author.}, Ping Hu$^{1}$, Xiaofeng Zhu$^{3}$, Na Zhao$^{2}$ 
}
\affiliations{
    $^{1}$School of Computer Science and Engineering,
    University of Electronic Science and Technology of China, Chengdu, China\\
    $^{2}$Information Systems Technology and Design Pillar,
    Singapore University of Technology and Design, Singapore\\
    $^{3}$School of Computer Science and Technology,
    Hainan University, Haikou, China
}
\usepackage{bibentry}

\begin{document}

\maketitle

\begin{abstract}
Graph-based methods have proven to be effective in capturing relationships among points for 3D point cloud analysis. However, these methods often suffer from suboptimal graph structures, particularly due to sparse connections at boundary points and noisy connections in junction areas. To address these challenges, we propose a novel method that integrates a graph smoothing module with an enhanced local geometry learning module. Specifically, we identify the limitations of conventional graph structures, particularly in handling boundary points and junction areas. In response, we introduce a graph smoothing module designed to optimize the graph structure and minimize the negative impact of unreliable sparse and noisy connections. Based on the optimized graph structure, we improve the feature extract function with local geometry information. These include shape features derived from adaptive geometric descriptors based on eigenvectors and distribution features obtained through cylindrical coordinate transformation.
Experimental results on real-world datasets validate the effectiveness of our method in various point cloud learning tasks, \ie classification, part segmentation, and semantic segmentation.
\end{abstract}

\begin{links}
    \link{Code}{https://github.com/shangboyuan/GSPoint}
\end{links}

\section{Introduction}\label{sec:intro}
Point clouds have become foundational to advanced 3D technologies, enabling critical applications in autonomous driving~\cite{sheng2022rethinking,sheng2025ct3d++,panimages}, robotic perception~\cite{wang2025uncertainty,zhao2024sdcot++,xiu2025geometric,jianggaussianblock,li2024end,han2024dual}, and 3D spatial reasoning~\cite{li2024laso,wangaffordbot,wang2025augrefer,yuan2025scene}. A central challenge in representing point clouds for these applications is the effective organization of inherently unstructured points into structured data formats.
This challenge stems from the absence of topological structure and irregular spatial distribution, which fundamentally preclude the direct application of conventional grid-based convolutional methods.
As a result, this structural incompatibility necessitates specialized approaches to extract meaningful hierarchical features while preserving geometric fidelity~\cite{Sohail2024,Wu2024}, motivating researchers to develop many specialized deep methods of point cloud analysis~\cite{wu2023sketch,wang2024fly}.

Previous deep methods of point cloud analysis can be partitioned into four subgroups,
\ie voxel-based methods, MLP-based methods, transformer-based methods, and graph-based methods.
Voxel-based methods~\cite{wu20153d,maturana2015voxnet,zhou2018voxelnet} convert point clouds into structured voxel grids and apply mature 3D convolutional models for effective feature extraction, but they ignore geometric precision to particularly affect the preservation of fine-grained details.
Instead of taking voxel as input, other methods use individual points as input.
Specifically,
MLP-based methods~\cite{Qi2017a,Qi2017,Thomas2019} avoid the information loss associated with grid conversion by directly operating on raw point clouds. However, raw point data often include noise to affect the effectiveness of feature extraction.
Transformer-based methods~\cite{Guo2021,Zhao2021,yu2022point} employ self-attention mechanisms to simultaneously focus on local details and global context, thereby enhancing the understanding of complex samples. However, they typically require a large number of data in the train process, placing high demands on computational resources.
Graph-based methods~\cite{Wang2019,Wang2019a,Du2022} explicitly model the relationships between points with graphs, where nodes represent individual points and edges encode spatial adjacency or feature affinities.
Graph can effectively organize the inherently unstructured points~\cite{mo2022simple,he2025graph}, and thus attracting much research interest.

Recent advances in graph-based methods have focused on two strategies, \ie fixed graph methods and graph learning methods.
Fixed graph methods capture the relationship between two points by the fixed graph structures with adaptive edge-weighting mechanisms to adjust convolution weights for neighboring points.
For instance, \citet{Wang2019} compute similarity weights based on position and feature disparities, while \citet{Xu2021} employ a weight bank combined with relative position to generate adaptive weights.
Graph learning methods use dynamic graph structures to capture multi-scale relationships.
For instance, \citet{yan2020pointasnl} propose to integrate attention-based global graph with a local graph constructed by the k-NN method, enabling the combined graph to capture global semantic relationships and local geometric structures.
\citet{Zhang2023} propose involving a correlation filter into a weighted local graph to suppress irrelevant connections from distant neighbors.
Dynamic graph structures have attracted growing attention in recent research, due to their flexibility in adjusting graph connections which can establish multiscale relationships and mitigate noise interference~\cite{peng2023grlc,Du2024,xu2024investigating,10833915}.

\begin{figure}[!t]
    \centering
    \setlength{\tabcolsep}{-1.1pt}
    \begin{subfigure}[b]{0.22\textwidth}
        \includegraphics[page=1,width=\textwidth]{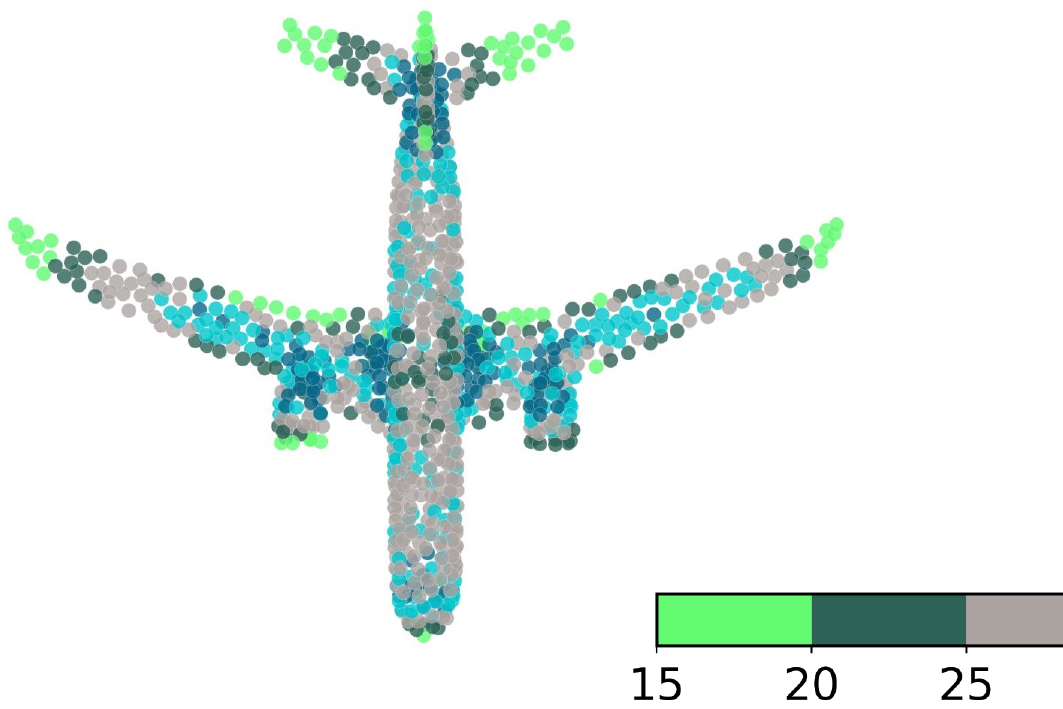}
        \caption{With normal ball query}
    \end{subfigure}
    \hspace{-0.21cm}
    \begin{subfigure}[b]{0.22\textwidth}
        \includegraphics[page=2,width=\textwidth]{heat_cut.pdf}
        \caption{With our method}
    \end{subfigure}
    \caption{
    Illustration of sparse connection.
    The heatmaps of degree distribution indicates that (a) When the graph is constructed using the normal ball query,
    some boundary points exhibit sparse connection and out-degree values below 20, whereas some points have out-degree values exceeding 35. (b) Through our method, the number of points with extreme out-degree values is reduced.}
    \label{fig:heat}
\end{figure}
\begin{figure}[!t]
    \begin{subfigure}[b]{0.22\textwidth}
        \includegraphics[width=\textwidth]{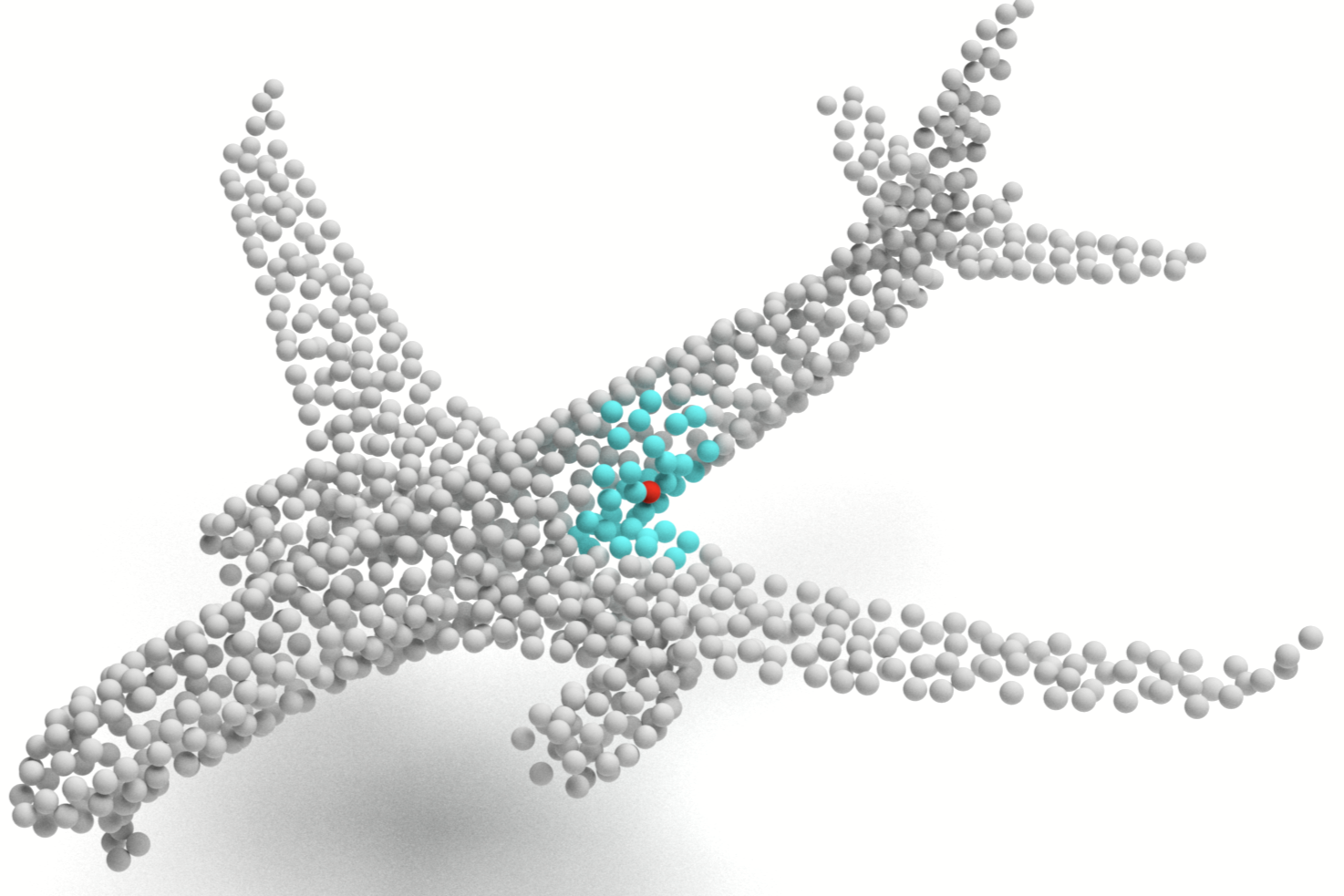}
        \caption{With normal ball query}
    \end{subfigure}
    \quad
    \begin{subfigure}[b]{0.22\textwidth}
        \includegraphics[width=\textwidth]{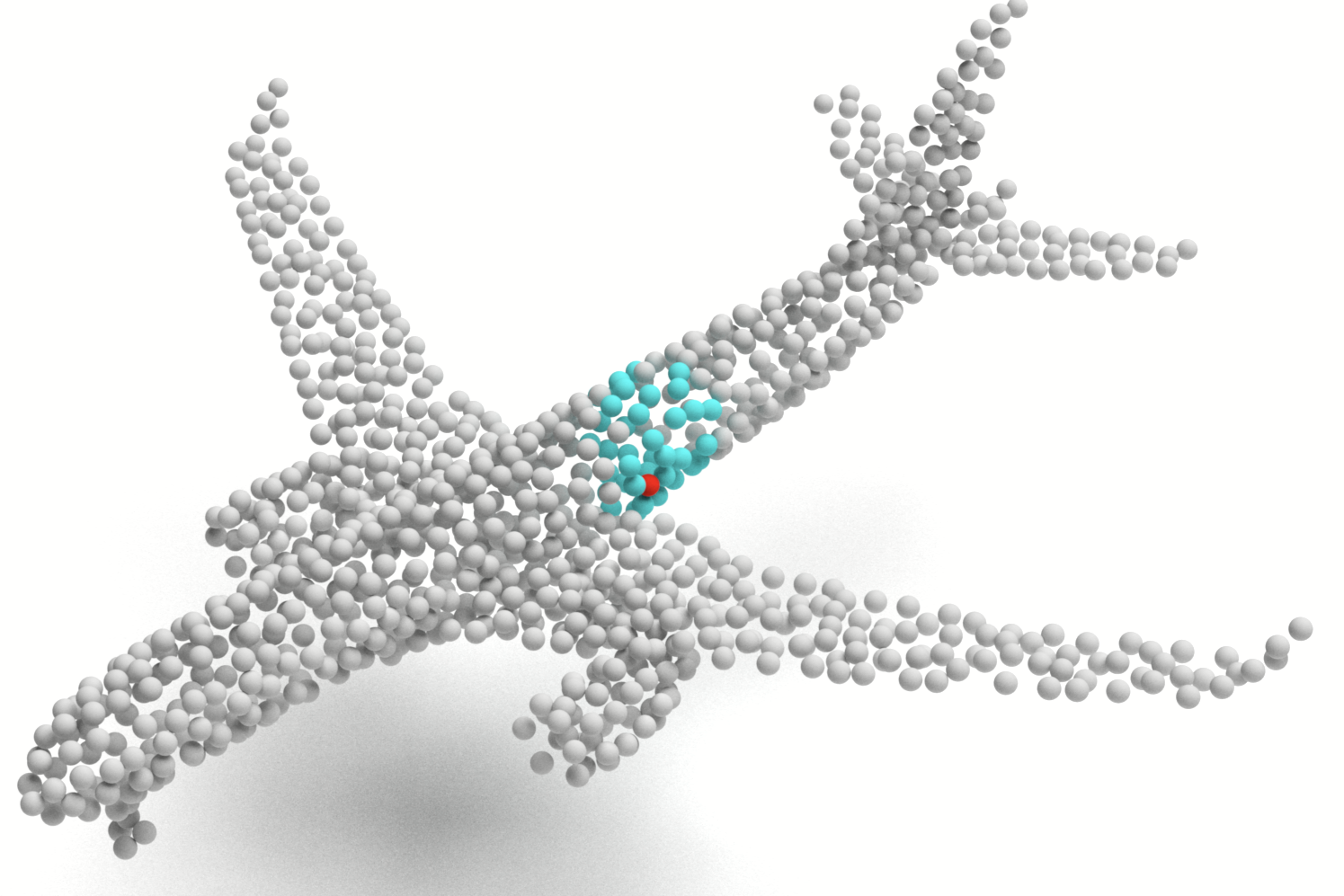}
        \caption{With our method}
    \end{subfigure}
    \caption{
    Illustration of noisy connection.
    The constructed point neighborhoods indicate that
    (a) By the normal ball query, the constructed point cloud neighborhood might exhibit noisy connection, \eg the neighbors of a point encompasses both fuselage and wing points.
    (b) Through our method, the neighbors of the point are refined to include only points from the fuselage.}
    \label{fig:neighbor}
\end{figure}

However, previous graph-based methods for 3D point cloud analysis still have limitations to be addressed.
First, sparse connections among boundary points hinder the effective usage of point cloud contour information.
As shown in Figure~\ref{fig:heat}(a), the out-degrees of boundary points are highly imbalanced due to the graph construction using a ball query. This causes their neighbors to be sparsely connected, making it difficult for useful boundary information to be sufficiently propagated in the representations of these points.
Second, noisy connections in junctions of different parts result in the unreliable point neighbor relationships.
As shown in Figure~\ref{fig:neighbor}(a), conventional graph construction methods usually rely on Euclidean distance metrics without adequately considering geometric variations, thereby leading to inaccurate and noisy neighbor modeling.

\begin{figure*}[ht]
    \centering
    \includegraphics[page=1,width=\linewidth]{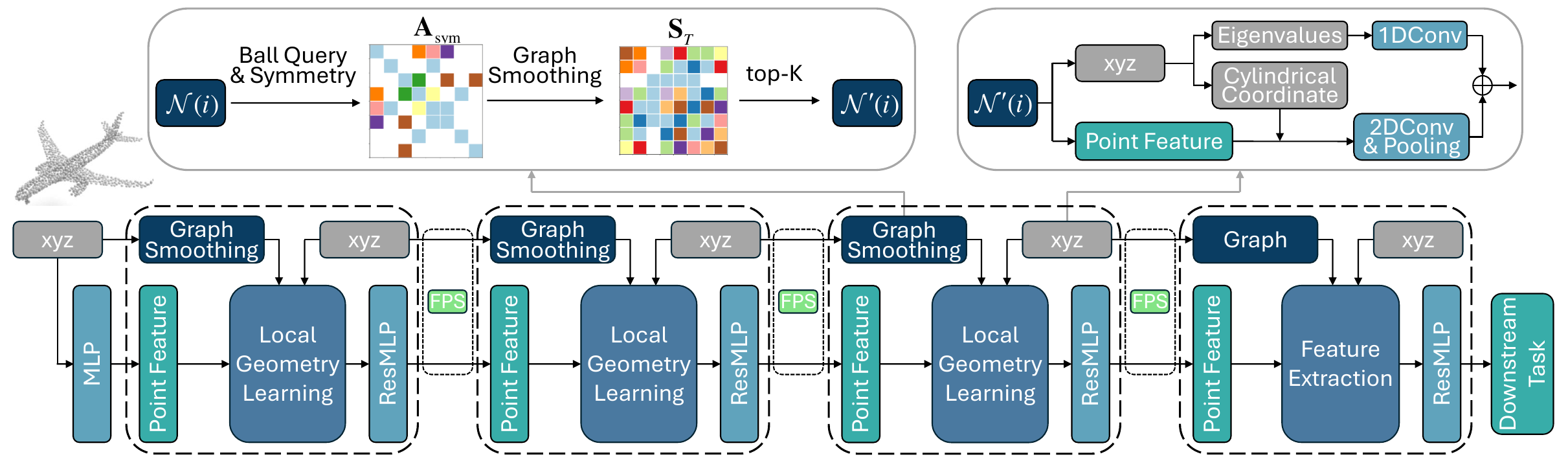}
    \caption{The proposed framework of GSPoint. Our method involves two key  parts in the point cloud feature extraction process, \ie graph smoothing module and local geometry learning module. Note that, FPS indicates farthest point sampling.}\label{fig:process}
\end{figure*}

In this paper, we propose a novel graph-based method equipped with \textbf{graph smoothing} and \textbf{local geometry learning} modules, which mitigate the two aforementioned issues, as demonstrated in Figures~\ref{fig:heat}(b) and \ref{fig:neighbor}(b).
To achieve this, we first systematically analyze the mechanisms responsible for the unreliable sparse and noisy connections originating from suboptimal graph structures, and then mitigate their side effects through a novel framework entitled GSPoint as in Figure~\ref{fig:process}.
Specifically, our graph smoothing module is established by balancing the disparities of degrees, where we optimize the graph to alleviate both sparse and noisy connections inherent in suboptimal graph structures, thereby constructing more robust neighbor relationships.
Subsequently, our local geometry learning module leverages local shape features and distribution features to provide geometric priors within the optimized point neighbors, thereby enhancing the feature extraction process.
These two modules are employed in multiple blocks with hierarchical downsampling strategy to achieve multi-scale feature extraction for downstream tasks.
Our contributions are summarized as follows:

\begin{itemize}
\item We analyze the limitations of conventional graph structures for boundary points and junction areas, and then propose a graph smoothing module in our method to optimize the graph structures for reducing the negative impact of unreliable sparse and noisy connections.
\item We introduce a local geometry learning module, where the adaptive shape features extract complex local geometric information and the cylindrical coordinate can capture spatial distribution relationships within the optimized neighborhood. These components help to extract discriminative features and promote downstream tasks.
\item Extensive experiments on benchmark datasets for both point cloud classification and segmentation tasks demonstrated that our method achieved competitive performance. Furthermore, ablation studies validated the individual contribution of each component in our method.
\end{itemize}


\section{Motivation and Analysis}\label{analysis}
To extract point cloud features that are effective for downstream tasks, conventional graph-based methods typically construct a graph to model the similarity between points and then leverage convolution operations to aggregate point cloud neighbor information. 
Given a set of a point cloud $\mathcal{P}=\{ \mathbf{p}_1, \mathbf{p}_2,..., \mathbf{p}_n\}$ where each point $\mathbf{p}_i \in \mathbb{R}^3$ contains 3D coordinates, the feature extraction function at the $l$-th stage can be formulated as follows:
\begin{equation}
\mathbf{x}_i^{(l+1)} = \mathcal{A} \Big( \sigma \Big( \psi 
\big( [\mathbf{x}_j^{(l)} \parallel (\mathbf{p}_i - \mathbf{p}_j) ] \big)_{j \in \mathcal{N}(i)} \Big) \Big), \\
\label{eq:feat_prop}
\end{equation}
where $\mathbf{x}_i^{(l)} \in \mathbb{R}^{\eta}$ denotes the feature vector of the $i$-th point at the $l$-th convolution stage, and $\parallel$ denotes vector concatenation. 
$\psi : \mathbb{R}^{\eta+3}\rightarrow \mathbb{R}^{\eta'}$ is a linear mapping function, 
$\sigma(\cdot)$ is an activation function, and $\mathcal{A}(\cdot)$ is a permutation-invariant aggregation operator (\eg, max-pooling).  

Here, $\mathcal{N}(i)$ denotes the neighbors of the point $\mathbf{p}_i$, and \textit{the construction of } $\mathcal{N}(i)$ \textit{directly affects the performance of graph-based methods}. Specifically,  $\mathcal{N}(i)$ is typically constructed using the ball query algorithm with a predefined query radius $r$ and a number of neighbors $k$, which can be mathematically represented as:
\begin{equation}
\mathcal{N}(i)=
\mathcal{B}_{r,k}(\mathbf{p}_i) = \{ \mathbf{p}_j \in \mathcal{P} \mid \rho_{ij} \le \min (\rho_{i(k)}, r) \},
\label{eq:ball_query_k}
\end{equation}
where $\rho_{i(k)}$ is the distance from point $\mathbf{p}_i$ to its $k$-th nearest neighbor, and $\rho_{ij}$ is the distance from $\mathbf{p}_i$ to $\mathbf{p}_j$.

Subsequently, we construct a sparsely connected graph $\mathcal{G} = (\mathcal{V}, \mathcal{E})$ based on the ball query method, where the vertices $\mathcal{V} = {1, 2, \dots, n}$ represent individual points, and the edges $\mathcal{E} \subseteq \mathcal{V} \times \mathcal{V}$ denote the connections between pairs of points. The edge between points $\mathbf{p}_i$ and $\mathbf{p}_j$ is defined as:
\begin{equation}
\mathcal{E} = \{ (i,j) \mid \mathbf{p}_j \in  \mathcal{B}_{r,k}(\mathbf{p}_i) \},
\label{eq:edges}
\end{equation}
and the corresponding adjacency matrix $\mathbf{A} \in \mathbb{R}^{n \times n}$ is a sparse matrix, where $a_{ij} =\{1, \text{if} (i,j) \in \mathcal{E};~ 0,\text{otherwise}\}. $

Previous work has primarily focused on improving the convolution process by adjusting the weights~\cite{Wang2019, Xu2021} or adding edges based on semantic similarity~\cite{yan2020pointasnl, Zhang2023} to $\mathcal{G}$, while overlooking the potential issue that the construction of the neighborhood $\mathcal{N}_k(i)$ could lead to a graph containing unreliable connections.
In particular, these unreliable connections in the graph are typically sparse at boundary points and noisy in junction areas, which limits information propagation and hinders the extraction of discriminative features.

Let $\mathcal{P} = \mathcal{P}_{\alpha} \cup \mathcal{P}_{\beta}$ denote the partition of the point cloud into points inside and outside the radius. That is, $\mathcal{P}_{\alpha}=\{\mathbf{p}_i \in \mathcal{P} \mid \rho_{i(k)} \leq r \}$ and 
$\mathcal{P}_{\beta}=\{\mathbf{p}_i \in \mathcal{P} \mid \rho_{i(k)} > r \}$. 

For $\mathbf{p}_u \in \mathcal{P}_{\alpha}$ and $\mathbf{p}_v \in \mathcal{P}_{\beta}$,
considering the constraints of radius $r$ and neighbor number $k$ in Eq.~(\ref{eq:ball_query_k}) we observe:
\begin{equation}
\rho_{u(k)} \leq r,; \rho_{v(k)} > r \quad \Rightarrow \quad r_u = \rho_{u(k)},; r_v = r
\label{eq:neq_radius}
\end{equation}
where $r_u$ and $r_v$ denote the actual radius of the two points. The variation of actual radius of different points indicates that the connections in the graph constructed by the ball query method are \textit{directional}. In other words, there exists:
\begin{equation}
\exists \rho_{i(k)} \leq \rho_{ij} < \rho_{j(k)}, \quad(i,j) \not\in \mathcal{E} ,\quad (j,i) \in \mathcal{E}.
\end{equation}

This directional property results in an inconsistency between the in-degree and out-degree.
To examine the degree relationships, we visualize the out-degree distribution for each point as a heatmap in Figure~\ref{fig:heat} and provide a comprehensive analysis below.

\noindent\textbf{Sparse Connection.}  
As shown in Figure~\ref{fig:heat}(a), boundary points are generally located at regions with high curvature or along structural edges, 
which often encapsulate key geometric features and semantic information. 
For boundary points, the spatial distribution of points around them is inherently sparse, resulting in fewer neighboring points within the search radius $r$ compared with internal points.
Accordingly, the out-degree of a boundary point is expressed as:
\begin{equation}
d_i^{\text{(out)}} \leq d_i^{\text{(in)}} \leq k,
\label{eq:d_bnd}
\end{equation}
where the out-degree $d^{\text{(out)}}_i$ indicates the number of neighbors that the point $\mathbf{p}_i$ connects to, while the in-degree $d^{\text{(in)}}_i$ indicates the number of points that include $\mathbf{p}_i$ as their neighbor. 
Due to the spatial distribution, fewer points include boundary points in their neighborhoods, resulting in sparse connections and this degree inequality.

These sparse connections restrict the propagation of boundary features to adjacent regions, potentially reducing the model's ability to extract discriminative geometric features crucial for shape characterization.

\noindent\textbf{Noisy Connection.} 
Junction points, which locate at the intersections of distinct instances, are characterized by high local point density where different instances exist in close Euclidean proximity. 
When employing ball query algorithms that rely solely on Euclidean distance metrics, these densely populated regions tend to generate noisy connections, as points from different instances are indiscriminately incorporated into the same neighborhood due to their spatial proximity.
To be specific, for a junction point $\mathbf{p}_i$, the degree inequality contributes to these noisy connections:
\begin{equation}
d_i^{\text{(out)}} \geq k = d_i^{\text{(in)}},
\label{eq:d_juk}
\end{equation}
with the in-degree $d^{(in)}_i$ equals exactly $k$ because a junction point can include exactly $k$ neighbors (as dictated by the algorithm’s constraint), while the out-degree $d^{out}_i$ exceeds $k$ due to the inclusion by multiple neighboring points. Such excessive and cross-instance connections, \ie noisy connections lead to a scenario in which junction points propagate their features across structurally distinct instances, thereby blurring the crucial geometric and semantic boundaries.

For example, as depicted in Figure~\ref{fig:neighbor}(a), a point on a fuselage might incorrectly incorporate several points from a wing into its neighborhood merely due to their spatial proximity, despite the clear structural separation.

\noindent\textbf{Our Motivation.} The analysis above demonstrates the correlation between the unreliable connections (sparse at boundaries and noisy at junctions) and the resulting imbalance in degree distributions within the graph. This imbalance inherently stems from the limitations of fixed graph structures that rely on the Euclidean-distance-based ball query algorithm. Consequently, these suboptimal graph structures limit the model's ability to extract discriminative features with the feature aggregation function as Eq.~(\ref{eq:feat_prop}).
In response, this paper aims to introduce a novel method which optimizes graph structures to address the unreliable sparse and noisy connections and incorporates richer geometric information to enable a more robust feature extraction.

\section{Method}
Our method is abbreviated as GSPoint whose framework is shown in Figure~\ref{fig:process}.
It employs a hierarchical downsampling architecture, where the graph smoothing module balances degree distribution through symmetric adjacency refinement and leverages graph smoothing process to capture multi-hop relationships within the graph.
The local geometry learning module exploits the enhanced geometric properties of within the optimized neighborhoods of the smoothed graph to extract more discriminative features.
Details are as follows.

\subsection{Graph Smoothing}
To address the limitations of unreliable sparse and noisy connections, 
we propose a graph smoothing module, where the symmetric adjacency refinement aims to establish balanced connections, and the finite steps graph smoothing aims to capture multi-hop relationships while maintain local similarities.
This approach achieves a more uniform degree distribution, \textit{cf.} Figure~\ref{fig:heat}(b), and facilitates a more robust neighborhood construction, \textit{cf.} Figure~\ref{fig:neighbor}(b), thereby enhancing the model's discriminative feature extraction capability.

\noindent\textbf{Symmetric Adjacency Refinement.} 
To resolve structural asymmetry, we first enforce symmetric connectivity constraints through floor operation:
\begin{equation}
\mathbf{A}_{\text{sym}} = \left\lfloor \frac{\mathbf{A} + \mathbf{A}^\top}{2} \right\rfloor.
\label{eq:symmetric}
\end{equation}

This process eliminates directional connections and creates a real symmetric matrix $\mathbf{A}_{\text{sym}}$.
After the transformation, the degrees satisfy the following relationship:
\begin{equation}
k\geq d_u^{(in)} = d_u^{(out)} \geq d_v^{(out)} = d_v^{(in)}, 
\label{eq:newdegree}
\end{equation}

This symmetry prevents imbalances in connections during the subsequent graph smoothing process. Furthermore, we leverage the degree inequality in Eq.(\ref{eq:newdegree}) for re-weighting $\mathbf{A}_{\text{sym}}$ and conduct symmetric normalization as follows:
\begin{equation}
\tilde{\mathbf{A}} = \mathbf{D}^{-1/2} \mathbf{A}_{\text{sym}} \mathbf{D}^{-1/2},
\label{eq:sym_normalize}
\end{equation}
where $\mathbf{D}$ is the degree matrix of $\mathbf{A}_{\text{sym}}$, and $d_i = \sum_{j=1}^n \tilde{a}_{ij}$ is the degree of node $i$ in the undirected graph.
The symmetrically normalized adjacency matrix $\tilde{\mathbf{A}}$ introduces normalized edge weight $\tilde{a}_{ij} = 1/{\sqrt{d_id_j}}$, inversely scaling with degrees.
This inverse proportionality underpins our \textbf{key insights}: \textit{low-degree boundary points gain higher weights}, while \textit{high-degree junction points are adaptively suppressed}, naturally balancing their influence in feature aggregation.

\noindent\textbf{Multi-hop Relationships in the Graph.} 
Direct utilization of the refined adjacency matrix is suboptimal as sparse and noisy connections remain. To further create more connections for boundary points and suppress noise, we consider multi-hop relationships in the graph. 

The weight between the $i$-th point and the $j$-th point considering all paths of length $T$ in the graph is given by: 
\begin{equation}
\begin{split}
(\tilde{\mathbf{A}}^T)_{ij} 
&=\sum_{\xi\in {\Xi}^T_{ij}}
\prod_{t=1}^T \frac{1}{\sqrt{d_{\xi_{t-1}} d_{\xi_t}}} ,
\label{eq:recursive}
\end{split}
\end{equation}
where $\xi_0=i$ and $\xi_T=j$, ${\Xi}^T_{ij}$ denotes the set of all possible paths of length $T$ and $\xi_t$ are intermediate points in the path.
Due to the property of degree distribution in the graph, we have $\frac{1}{\sqrt{d_v}}\cdot\frac{1}{\sqrt{d_{\xi_1}}}>\frac{1}{\sqrt{d_u}}\cdot\frac{1}{\sqrt{d_{\xi_1}}}$ for a point $\mathbf{p}_v$ with sparse connections and a point $\mathbf{p}_u$ with more connections. When comparing paths with identical intermediate points $(\xi_1,\xi_2,..,\xi_{T-1})$ leading to $\mathbf{p}_j$, the path starting from $\mathbf{p}_v$ to $\mathbf{p}_j$ exceeds that from $\mathbf{p}_u$ to $\mathbf{p}_j$, \ie
\begin{equation}
(\tilde{\mathbf{A}}^T)_{vj}  > (\tilde{\mathbf{A}}^T)_{uj},
\end{equation}
which indicates that within the $T$-hop neighborhood, the lower-degree point $\mathbf{p}_v$ has higher propagation weights $(\tilde{\mathbf{A}}^T)_{vj}$ compared to the higher-degree point $\mathbf{p}_u$.
This is expected to facilitate more connections to boundary points while mitigating the noise from junction points.

\noindent\textbf{von Neumann Kernel.} 
However, the high-order terms of $\tilde{\mathbf{A}}$ present two challenges. 
First, $\tilde{\mathbf{A}}^T$ is numerically unstable as the order $T$ increases. Second, $\tilde{\mathbf{A}}^T$ cannot maintain local consistency, as it only considers paths exactly equal to $T$ in length, ignoring shorter paths that are critical for the consistency of the local structure. 
Based on these considerations, we introduce the von Neumann kernel~\cite{neumann2016propagation} as the smoothing kernel in our graph smoothing process, which can be expressed as:
\begin{equation}
K_\text{{NEU}}
= (I-\alpha \tilde{\mathbf{A}})^{-1}
= \lim_{T\rightarrow\infty} \sum_{t=0}^{T} (\alpha \tilde{\mathbf{A}})^t,
\label{eq:lim}
\end{equation}
where $I$ denotes the identity matrix, and $\alpha \in (0,1)$ is an attenuation factor that controls the trade-off between local consistency and global connectivity by progressively suppressing higher-order terms. 

The matrix $K_\text{{NEU}}$ can be viewed as a kernel function over the vertices of the graph, where $K_\text{{NEU}}(i,j)$ represents the $(i,j)$-th element of the matrix, indicating the strength of connection between vertices $i$ and $j$ after smoothing. The positive definiteness of $K_\text{{NEU}}$ ensures that:
\begin{equation}
\forall  i\neq j,\quad K_{\text{NEU}}(i,i) \geq K_{\text{NEU}}(i,j),
\label{eq:posdef}
\end{equation}
which guarantees that each point maintains maximum weights on itself during the smoothing process. Compared with $\tilde{\mathbf{A}}^T$ which only considers paths of exact length of $T$, the von Neumann kernel considers all paths from 1 to $T$, providing a more stable graph structure.

\noindent\textbf{Finite Steps Graph Smoothing Process.} 
For practical implementation of the von Neumann kernel, we approximate it with a finite sum to define our multi-hop graph smoothing process with the finite smoothing order $T$:   
\begin{equation}
\mathbf{S}_T = \sum_{t=0}^{T} (\alpha \tilde{\mathbf{A}})^t,\quad \alpha \in (0,1),
\label{eq:Sn}
\end{equation}
and we obtain the final neighborhoods $\mathcal{N}'(i)$ through the top-K selection for each row of $\mathbf{S}_T$. This process preserves the boundary amplification effect while maintaining computational efficiency. The progressive summation over $t=0,...,T$ optimizes the neighborhood of each point where sparsely connected boundary points rank higher in the top-K selection, with simultaneous suppression of noisy connections from junction points.

\subsection{Local Geometry Learning}
To exploit the geometric representation of each point’s local neighborhood, which has been optimized by our graph smoothing method, we enhance the feature extraction function with two sets of adaptive features, \ie local shape features and local distribution features.

\noindent\textbf{Local Shape Features.}
For a set of points, the eigenvalues of its covariance matrix contain rich geometry information. Previous works~\cite{dong2017selection,lin2014eigen} derive classical shape descriptors such as planarity, sphericity, and linearity as hand‑crafted geometric features. 
However, fixed descriptors might be not adaptable in complex structures or scenes.
Therefore, we use a learnable network to transfer the eigenvalues into adaptive shape features.

Specifically, for each point $\mathbf{p}_i$ with its $\mathcal{N}'(i)$, we
perform eigenvalue decomposition on its covariance matrix $\mathbf{C}_i$:
\begin{equation}
\mathbf{C}_i = \mathbf{V}_i \mathbf{\Lambda}_i \mathbf{V}_i^\top,
\label{decom}
\end{equation}
where $\mathbf{\Lambda}_i=\text{diag}(\lambda_i^{(1)}, \lambda_i^{(2)}, \lambda_i^{(3)})$ contains the eigenvalues $\lambda_i^{(1)} \geq \lambda_i^{(2)} \geq \lambda_i^{(3)}\geq 0$, and $\mathbf{V}_i=[\mathbf{v}_i^{(1)},\mathbf{v}_i^{(2)},\mathbf{v}_i^{(3)}]$ consists of the corresponding orthonormal eigenvectors of $\mathbf{C}_i$.

Then we feed the eigenvalues into a learnable MLP network projecting them into $\eta'$-dimensional feature space as local shape features and denote it as $\phi(\mathbf{\Lambda})$.

\noindent\textbf{Local Distribution Features.} 
To further capture the anisotropic and distance distribution information of the optimized neighborhoods, we complement the local geometry features with a cylindrical coordinate transformation that aligns the local structure along its principal axes.
Based on the above eigenvalue decomposition, we compute the cylindrical coordinates of each neighbor point relative to the query point by projecting the displacement vector $\Delta\mathbf{p}_j=\mathbf{p}_j-\mathbf{p}_i$ onto three principal axes: 
\begin{equation}
(x'_j,\,y'_j,\,z'_j)
=
\bigl(\Delta\mathbf{p}_j\!\cdot\!\mathbf{v}_i^{(2)},\,
      \Delta\mathbf{p}_j\!\cdot\!\mathbf{v}_i^{(3)},\,
      \Delta\mathbf{p}_j\!\cdot\!\mathbf{v}_i^{(1)}\bigr),
\label{newxyz}
\end{equation}
and the new coordinate system is given by:
\begin{equation}
\bigl(h_j,\,\omega_j,\,\cos\theta_j\bigr)
=
\Bigl(z'_j,\,
       \sqrt{{x'_j}^2+{y'_j}^2},\,
       {x'_j}/{\sqrt{{x'_j}^2+{y'_j}^2}}\Bigr).
 \label{eq:h}
\end{equation}

To ensure numerical stability, both the height and radial distance are divided by their respective maximum values in the neighborhood, yielding the normalized values $h'$ and $\omega'$.

Then, the neighbor points in the original Cartesian coordinates are transformed by $(x,y,z) \rightarrow (h',\omega',\cos \theta)$ into the new cylindrical coordinate system as $\mathbf{p'} = (h',\omega',\cos \theta)$, where $h'$ quantifies the axial anisotropy and $\omega'$ characterizes the radial distance distribution. 
This transformation enables us to leverage the local distribution information to implement our feature extraction function.

Finally, our feature extraction function with the mapping function $\psi' : \mathbb{R}^{\eta+6}\rightarrow \mathbb{R}^{\eta'}$ can be formulated as:
\begin{align}
&\mathbf{x}_i^{(l+1)} 
= \notag \\
&\mathcal{A} \Big( \sigma \Big( \psi' 
\big( [\mathbf{x}_j^{(l)}  \parallel (\mathbf{p}_i - \mathbf{p}_j) \parallel  \mathbf{p'}_j^{(l)}] \big)_{j \in \mathcal{N}'(i)} \Big) \Big) \parallel \phi( \mathbf{\Lambda}_i).
\label{eq:fuse}
\end{align}

We integrate this enhanced  feature extraction function with our graph smoothing module and hierarchical downsampling strategies, as illustrated in Figure~\ref{fig:process}, to achieve multi-stage feature extraction for point cloud analysis.

\section{Experiments}

\noindent\textbf{Comparison Methods.}
For point cloud classification and segmentation tasks, we compare our method with the following representative
point cloud analysis methods:
PointNet++~\cite{Qi2017a}, PointTrans.~\cite{Zhao2021}, PointMLP~\cite{Ma2022}, PointNeXt~\cite{Qian2022}, PointMAE~\cite{pang2022masked}, PointGPT~\cite{chen2023pointgpt}, GSLCN~\cite{Liang2023}, PointWavelet~\cite{wen2024pointwavelet} and DuGREAT~\cite{li2025dual}.

\subsection{Point Cloud Classification}
We conduct point cloud classification tasks on ModelNet40~\cite{wu20153d} and ScanObjectNN~\cite{uy2019revisiting}.

\noindent\textbf{Experimental Setups.}
The ModelNet40 dataset is a benchmark for synthetic 3D object classification. It comprises 9,843 training samples and 2,468 testing samples across 40 object categories. 
The ScanObjectNN dataset consists of 15,000 point cloud samples extracted from 2,902 unique object instances spanning 15 categories. In particular, the PB\_T50\_RS subset of ScanObjectNN is particularly challenging due to real-world scanning artifacts such as noise, occlusion, and rotations.
Following previous work \cite{Qian2022,Ma2022}, we use 1024 points without normals as input for both datasets. For ModelNet40, we apply random translations and train the model for 500 epochs. 
For ScanObjectNN, we augment data with random scaling and rotations, and train the model for 250 epochs.

\noindent\textbf{Classification on ModelNet40 and ScanObjectNN.}
In Table~\ref{tab:M40nSobj}, we report the results with mean accuracy (mAcc) and overall accuracy (OA) to evaluate the effectiveness of all comparison method on object-level point cloud classification tasks.
For ModelNet40, our method achieves an mAcc of 91.5\% and an OA of 94.5\%, demonstrating its competitive performance on synthetic object-level point cloud classification tasks.
On the PB\_T50\_RS subset of the ScanObjectNN dataset, our method obtains a mAcc of 86.4\% and an OA of 88.1\%.
These results validate the good generalization of our method under challenging real-world conditions, including adverse factors of noise, occlusion, and rotation.

\begin{table}[!t]
\small
\centering
\renewcommand\tabcolsep{4.0pt} 
\resizebox{84mm}{!}{
\begin{tabular}{lcccc}
\toprule
\multirow{2}{*}{Method} & \multicolumn{2}{c}{ModelNet40} & \multicolumn{2}{c}{ScanObjectNN}\\ 
& mAcc(\%) & OA(\%) & mAcc(\%) & OA(\%)\\
\hline
PointNet++ (2017) & 88.5& 91.9& 69.8& 73.7\\
PointTrans. (2021) &  90.6 & 93.7 & --& --\\
PointMLP (2022)& 91.3&94.1& 83.9& 85.4\\
PointNeXt (2022)& 90.8&93.2& 85.8& 87.7\\
PointMAE (2022) & --& 93.8& --& 85.2\\
PointGPT-S (2023) & --& 94.0& --& 86.9\\
GSLCN (2023)& 91.4& 94.2& 84.1&85.8\\
PointWavelet (2024)& 91.1& 94.3& 85.8& 87.7\\
DuGREAT (2025) & 90.9 & 94.0 & 84.5& 87.1\\
\hline
GSPoint (ours)& \textbf{91.5}& \textbf{94.5}& \textbf{86.4}& \textbf{88.1}\\
\bottomrule
\end{tabular}}
\caption{Classification on ModelNet40 and ScanObjectNN}\label{tab:M40nSobj}
\end{table}

\begin{table}[!t]
\small
\centering
\renewcommand\tabcolsep{4.5pt} 
\resizebox{84mm}{!}{
\begin{tabular}{lcc}
\toprule
Method & ~~Cls.mIoU (\%)~~& ~~Ins.mIoU (\%)~~\\
\hline
PointNet++ (2017) & 81.9& 85.1 \\
PointMLP (2022) & 84.6& 86.1 \\
PointNeXt (2022) & 85.2& 87.0 \\
GSLCN (2023)& 85.4&87.1\\
PointWavelet (2024) & 85.2& 86.8 \\
DuGREAT (2025)& 84.9&86.5\\
\hline
GSPoint (ours) & \textbf{85.6}& \textbf{87.2} \\
\bottomrule
\end{tabular}}
\caption{Part Segmentation on ShapeNetPart}\label{tab:shapenet}
\end{table}

\begin{table}[!t]
\small
\centering
\renewcommand\tabcolsep{5pt} 
\resizebox{84mm}{!}{
\begin{tabular}{l c c}
\toprule
Method & ~~~~mIoU (\%)~~~~~ & ~~~~OA (\%)~~~~~  \\
\hline
PointNet++ (2017) & 56.0 & 86.4 \\
PointNeXt (2022)& 70.5 & 90.6\\
GSLCN (2023)& 68.1&90.5\\
PointWavelet (2024)& 71.3 & --\\
\hline
GSPoint (ours) & \textbf{71.5} & \textbf{91.2} \\
\bottomrule
\end{tabular}}
\caption{Indoor Scene Segmentation on S3DIS}\label{tab:s3dis_comparison}
\end{table}

\subsection{Point Cloud Segmentation}
We further evaluate our method on two challenging benchmarks for point cloud segmentation: part segmentation on ShapeNetPart~\cite{chang2015shapenet} and indoor scene segmentation on S3DIS~\cite{armeni20163d}.

\noindent\textbf{Experimental Setups.}
The ShapeNetPart dataset is widely used for 3D object part segmentation. It comprises 12,137 training models and 2,874 testing models across 16 object categories, with each model annotated with 2 to 6 parts, resulting in a total of 50 distinct part labels.  
The S3DIS dataset includes six large-scale indoor areas with 271 rooms, where each point is annotated with one of 13 semantic categories, enabling rigorous studies in 3D semantic segmentation.
Typically, Area5 is reserved for testing the generalization performance on unseen indoor scenes.
For the ShapeNetPart dataset, each 3D shape is uniformly sampled to 2048 points. The training process spans 300 epochs, and data augmentation strategies including random scaling and point cloud jittering are applied.  
Following standard practice in previous work \cite{Thomas2019}, the raw point clouds from the S3DIS dataset are downsampled using voxels with a size of 0.04 m. Then the model is trained for 100 epochs using a combination of random scaling, rotation, and jittering.

\noindent\textbf{Part Segmentation on ShapeNetPart.}
In Table~\ref{tab:shapenet}, we report the results with classification mean IoU (Cls.mIoU) and instance mean IoU (Ins.mIoU) to evaluate the effectiveness of all comparison method on point cloud part segmentation tasks on ShapeNetPart.
Our approach achieves a Cls.mIoU of 85.6\% and an Ins.mIoU of 87.2\%, indicating that our proposed method is sufficient to preserve the local geometric details of object parts.

\noindent\textbf{Indoor Scene Segmentation on S3DIS Area5.}
In Table~\ref{tab:s3dis_comparison}, we report the results with mean IoU (mIoU), mean accuracy (mAcc) and overall accuracy (OA) to evaluate the effectiveness of indoor scene segmentation task on the S3DIS Area5 benchmark.  
Our method achieves 71.5\% mIoU, 77.8\% mAcc, and 91.2\% OA, demonstrating the robustness of our method within indoor scenes.

\subsection{Ablation Study}
\begin{table}[!t]
\small
\centering
\renewcommand\tabcolsep{3.0pt} 
\resizebox{84mm}{!}{
\begin{tabular}{c|cccc|cccc}
\hline
\multirow{2}*{}&\multicolumn{4}{c|}{Component} & \multicolumn{4}{c}{Dataset (metric)}   \\ \hline
&\multirow{3}*{SA}& 
\multirow{3}*{GS}& 
\multirow{3}*{$\mathbf{\Lambda}$}&
\multirow{3}*{$\mathbf{p}'$}&
Model-& Scan-& Shape-& S3DIS\\
 & & & & & Net40& ObjectNN &NetPart & Area5\\
 & & & & & (OA)&(OA) &(Ins.mIoU) &(mIoU) \\\hline
A&& & & & 92.6&86.9&86.5&68.2\\\hline
B&$\checkmark$ & & & & 92.6 &85.2 &85.4 &67.4\\
C&& $\checkmark$ & & & 93.6& 86.6 &86.9&68.3\\
D&$\checkmark$ & $\checkmark$ & & & 93.9&87.3 &87.0 &70.2\\ \hline
E& & & $\checkmark$ & & 93.4& 87.0& 86.7&69.2\\
F& & & & $\checkmark$ & 93.6& 87.1& 86.6&69.6\\
G& & & $\checkmark$ & $\checkmark$ & 93.6& 87.1& 86.9&69.8\\\hline
H&$\checkmark$& $\checkmark$& $\checkmark$& & 94.0& 87.5 &87.1 &70.9\\
I&$\checkmark$& $\checkmark$& & $\checkmark$& 94.3& 87.9 &87.1 &70.4\\\hline
J&$\checkmark$ & $\checkmark$ & $\checkmark$ & $\checkmark$& \textbf{94.5}&\textbf{88.1} &\textbf{87.2} &\textbf{71.5}\\
\hline
\end{tabular}}
\caption{Ablation study on our method's key components. SA: symmetric adjacency refinement, GS: graph smoothing, $\mathbf{\Lambda}$: adaptive shape features, $\mathbf{p}'$: cylindrical coordinates.}\label{tab:ablation}
\end{table}

We perform ablation study to verify the components, and visualization to better understand how our method works.

\noindent\textbf{Effectiveness of Different Components.}
As shown in Table~\ref{tab:ablation}, we evaluate the contributions of key components through ten experimental configurations.
Item-A reports the results of our model variant without the proposed four components.
From Items-B,C,D, we can observe that symmetric adjacency refinement (SA)  and graph smoothing (GS) alone play a small role, while the synergistic combination of SA and GS yields consistent improvements across all benchmarks (+1.3\%, +0.4\%, +0.5\%, +2.0\% improvements on ModelNet40, ScanObjectNN, ShapeNetPart, S3DIS).
From Items-E,F,G, the similar results can be observed from the combination of using adaptive shape features ($\mathbf{\Lambda}$) and cylindrical coordinates ($\mathbf{p}'$).
Compared with Item-A, it has +1.3\%, +0.2\%, +0.6\%, +2.2\% improvements on ModelNet40, ScanObjectNN, ShapeNetPart, S3DIS.
Subsequent incorporation of (SA, GS, $\mathbf{\Lambda}$) in Item-H and (SA, GS, $\mathbf{p}'$) in Item-I further boosts performance, validating the effectiveness of encoding geometric priors in feature aggregation combined with the graph smoothing module.
The full configuration (Item-J) achieves optimal results through component collaboration, 
confirming the effectiveness of our proposed framework across diverse tasks.

\noindent\textbf{Effectiveness of Graph Smoothing as a Plug-in.}
We conducted the experiments by integrating our graph smoothing method into other point cloud analysis methods in Table~\ref{tab:plugin}. As shown, our method consistently improves the performance on multiple tasks across different methods, further demonstrating its effectiveness and general applicability.

\begin{table}[!t]
\small
\centering
\renewcommand\tabcolsep{3.0pt} 
\resizebox{84mm}{!}{
\begin{tabular}{lcccc}
\toprule
\multirow{3}*{Ours+Method}&Model-& Scan-& Shape-& S3DIS\\
 &Net40& ObjectNN &NetPart & Area5\\
 &(OA)&(OA) &(Ins.mIoU) &(mIoU) \\
\hline
+~PointNet++ & 93.0($\uparrow$1.1) & 82.9($\uparrow$9.2) & 84.1($\uparrow$2.2) & 63.9($\uparrow$7.9) \\
+~PointMLP   & 94.4($\uparrow$0.3) & 85.8($\uparrow$0.4) & 85.0($\uparrow$0.4) & -- \\
+~PointNeXt  & 93.8($\uparrow$0.6) & 87.9($\uparrow$0.2) & 85.4($\uparrow$0.2) & 70.8($\uparrow$0.3) \\
\bottomrule
\end{tabular}}
\caption{The performance gain of our graph smoothing method as a plug-in for other point cloud analysis methods.} \label{tab:plugin}
\end{table}

\begin{figure}[!t]
\centering
    \begin{subfigure}[b]{0.15\textwidth}
        \includegraphics[page=1, width=\textwidth]{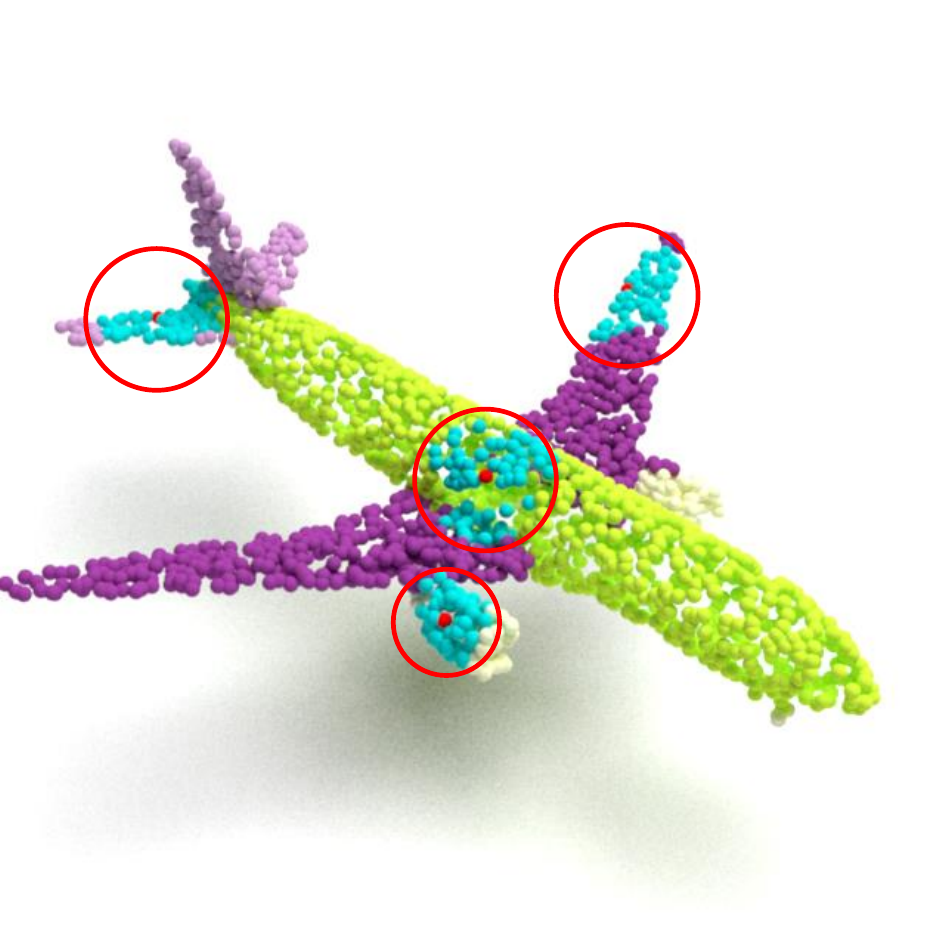}
    \end{subfigure}
    \begin{subfigure}[b]{0.15\textwidth}
        \includegraphics[page=3, width=\textwidth]{shapenet.pdf}
    \end{subfigure}
        \begin{subfigure}[b]{0.15\textwidth}
        \includegraphics[page=5, width=\textwidth]{shapenet.pdf}
    \end{subfigure}
        \begin{subfigure}[b]{0.15\textwidth}
        \includegraphics[page=2, width=\textwidth]{shapenet.pdf}
    \end{subfigure}
        \begin{subfigure}[b]{0.15\textwidth}
        \includegraphics[page=4, width=\textwidth]{shapenet.pdf}
    \end{subfigure}
        \begin{subfigure}[b]{0.15\textwidth}
        \includegraphics[page=6, width=\textwidth]{shapenet.pdf}
    \end{subfigure}
    \caption{With normal ball query (upper row) $vs.$ With our graph smoothing (bottom row).}\label{fig:ner}
\end{figure}

\begin{figure}[!t]
\centering
    \begin{subfigure}[b]{0.23\textwidth}
        \includegraphics[page=2, width=\textwidth]{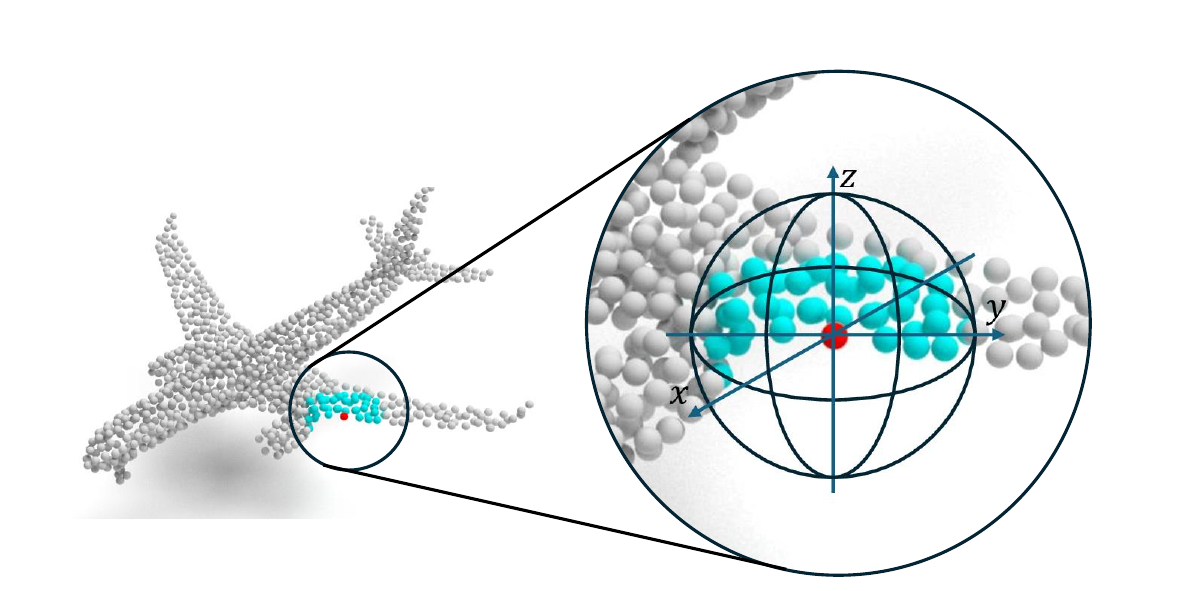}
        \caption{With xyz coordinate}
    \end{subfigure}
    \begin{subfigure}[b]{0.23\textwidth}
        \includegraphics[page=3, width=\textwidth]{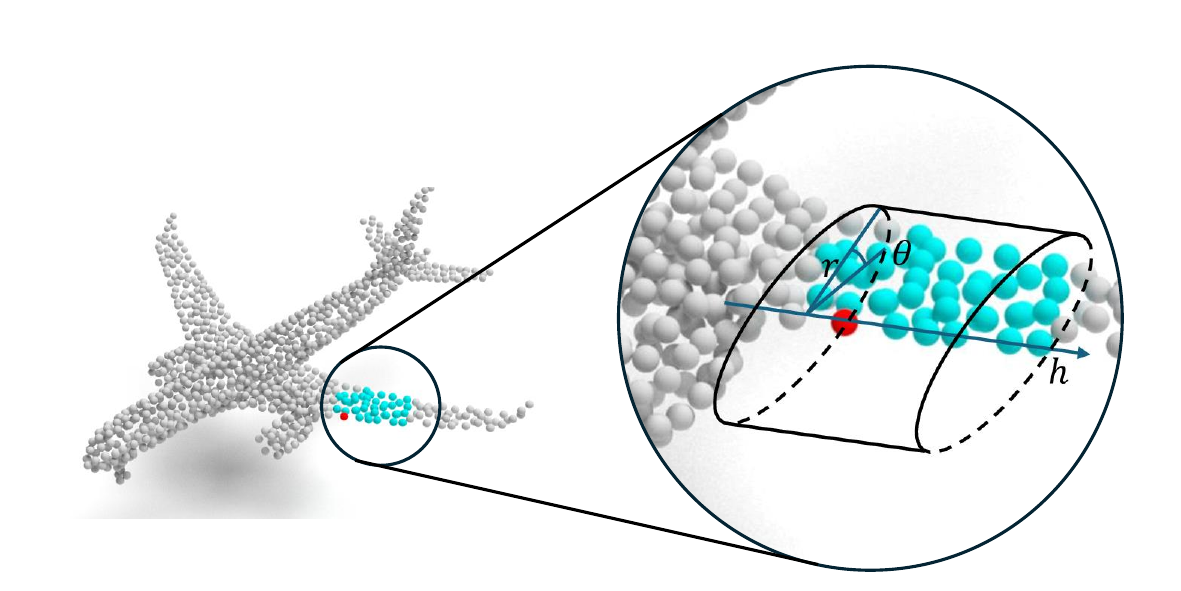}
        \caption{With cylindrical coordinate}
    \end{subfigure}
    \caption{xyz coordinate $vs.$ cylindrical coordinate.}
    \label{fig:xyz}
\end{figure}

\noindent\textbf{Visualization Validation.}
Figure~\ref{fig:ner} shows the neighbors of the points obtained by our proposed graph smoothing method versus the neighbors of the points obtained by the normal ball query method. By comparing the details, we observe that our method constructs more accurate neighbors for specific points, effectively mitigates the sparse and noisy connections introduced by previous graph construction strategy. 
Figure~\ref{fig:xyz} shows that in the process of our local geometry learning, the neighbors of the points obtained by our proposed cylindrical coordinate transformation and the neighbors of the points obtained by ordinary three-dimensional coordinates.
It is also observed that our method can capture the geometric information of objects well and avoid the limitations of Euclidean space measurement.

\begin{figure}[!t]
\centering
    \begin{subfigure}[b]{0.2343\textwidth}
        \includegraphics[page=1, width=\textwidth]{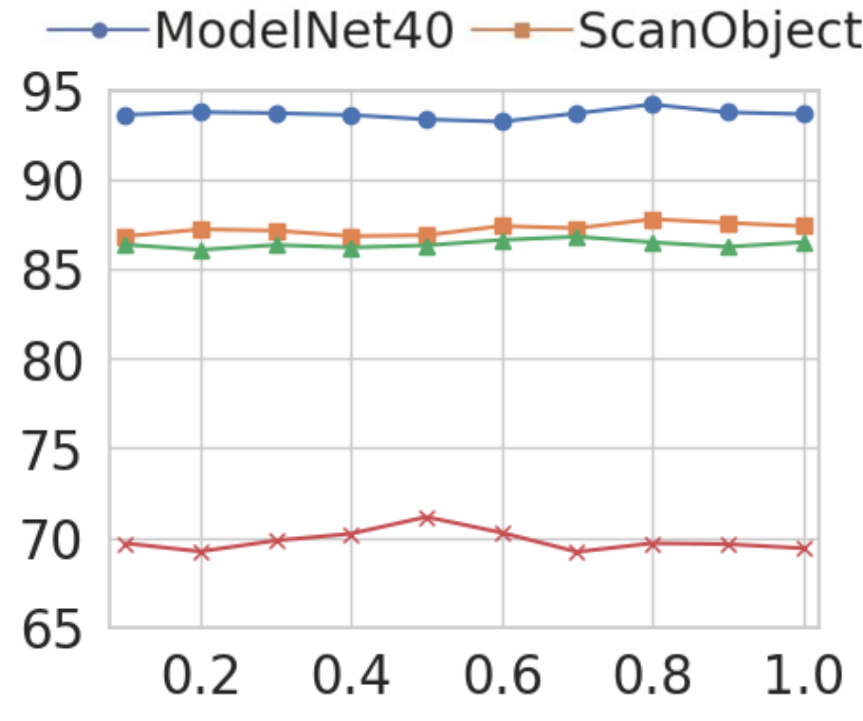}
        \caption{The attenuation factor $\alpha$}
    \end{subfigure}
    \begin{subfigure}[b]{0.2343\textwidth}
        \includegraphics[page=2, width=\textwidth]{abla.pdf}
        \caption{The smoothing order $T$}
    \end{subfigure}
    \caption{Hyper-parameter analysis.}
    \label{hp}
\end{figure}
\noindent\textbf{Hyper-Parameter Analysis.} The main hyper-parameters in our method include $\alpha$ as the attenuation factor and $T$ as the smoothing order.
By iterating over their range of values with other fixed settings, we get the corresponding evaluation metrics OA on the ModelNet40 and ScanObjectNN datasets, Ins.mIoU on ShapeNetPart dataset and mIoU on S3DIS dataset.
The results are shown in Figure~\ref{hp}, we can easily observe that
on datasets ModelNet40, ScanObjectNN and ShapeNetPart, both hyper-parameters $\alpha$ and $T$ are insensitive despite facing different classification and segmentation tasks.
For the semantic segmentation task on the dataset S3DIS, our method performs better on the interval $[0.4,0.6]$ of $\alpha$ and the interval $[3,4]$ of $T$.

\section{Conclusion}
In this paper, we provide an insightful analysis of traditional graph construction and ball query methods, identifying their inherent issue of suboptimal graph structures, which result in sparse connections at boundary points and noisy connections in junction areas. We propose a novel approach with a graph smoothing module to refine the structure and a local geometry learning module that enhances feature learning using adaptive geometric features and cylindrical coordinate transformation. Extensive experiments demonstrate the effectiveness of our method in classification and segmentation tasks.
Future work will focus on improving model efficiency and generalization to unseen categories through self-supervised learning, further advancing the practical applications of graph-based point cloud learning.

\section{Acknowledgments}
This research was supported in part by the National Key Research \& Development Program of China under Grant 2022YFA1004100,
in part by National Natural Science Foundation of China under Grant 62476048,
and in part by the Ministry of Education, Singapore, under its MOE Academic Research Fund Tier 2 (MOE-T2EP20124-0013).

\bibliography{aaai2026.bib}

@String{Computing = "Computing" }

@String{Computer = "{IEEE} Computer" }

@InProceedings{Qi2017,
  author    = {Qi, Charles R and Su, Hao and Mo, Kaichun and Guibas, Leonidas J},
  booktitle = {Proceedings of the IEEE/CVF Conference on Computer Vision and Pattern Recognition},
  title     = {Pointnet: Deep learning on point sets for 3d classification and segmentation},
  year      = {2017},
  pages     = {652--660},
}

@inproceedings{zhou2018voxelnet,
  title={Voxelnet: End-to-end learning for point cloud based 3d object detection},
  author={Zhou, Yin and Tuzel, Oncel},
  booktitle={Proceedings of the IEEE/CVF Conference on Computer Vision and Pattern Recognition},
  pages={4490--4499},
  year={2018}
}

@inproceedings{maturana2015voxnet,
  title={Voxnet: A 3d convolutional neural network for real-time object recognition},
  author={Maturana, Daniel and Scherer, Sebastian},
  booktitle={IEEE/RSJ International Conference on Intelligent Robots and Systems},
  pages={922--928},
  year={2015}
}

@Article{Du2022,
  author    = {Du, Zijin and Ye, Hailiang and Cao, Feilong},
  journal   = {IEEE Transactions on Neural Networks and Learning Systems},
  title     = {A novel local--global graph convolutional method for point cloud semantic segmentation},
  year      = {2022},
  number    = {4},
  pages     = {4798--4812},
  volume    = {35},
}

@Article{Guo2021,
  author    = {Guo, Meng-Hao and Cai, Jun-Xiong and Liu, Zheng-Ning and Mu, Tai-Jiang and Martin, Ralph R and Hu, Shi-Min},
  journal   = {Computational Visual Media},
  title     = {Pct: Point cloud transformer},
  year      = {2021},
  pages     = {187--199},
  volume    = {7}
}

@article{neumann2016propagation,
  title={Propagation kernels: efficient graph kernels from propagated information},
  author={Neumann, Marion and Garnett, Roman and Bauckhage, Christian and Kersting, Kristian},
  journal={Machine Learning},
  volume={102},
  pages={209--245},
  year={2016}
}

@article{dong2017selection,
  title={Selection of LiDAR geometric features with adaptive neighborhood size for urban land cover classification},
  author={Dong, Weihua and Lan, Jianhang and Liang, Shunlin and Yao, Wei and Zhan, Zhicheng},
  journal={International Journal of Applied Earth Observation and Geoinformation},
  volume={60},
  pages={99--110},
  year={2017}
}

@article{li2025dual,
  title={Dual-path geometric relation-aware transformer for point cloud classification and segmentation},
  author={Li, Xiangli and Wang, Qifan and Qiu, Baozhi},
  journal={Applied Soft Computing},
  volume={174},
  pages={112801},
  year={2025}
}

@Article{Ma2022,
  author  = {Ma, Xu and Qin, Can and You, Haoxuan and Ran, Haoxi and Fu, Yun},
  journal = {arXiv preprint arXiv:2202.07123},
  title   = {Rethinking network design and local geometry in point cloud: A simple residual MLP framework},
  year    = {2022}
}

@article{lin2014eigen,
  title={Eigen-feature analysis of weighted covariance matrices for LiDAR point cloud classification},
  author={Lin, Chao-Hung and Chen, Jyun-Yuan and Su, Po-Lin and Chen, Chung-Hao},
  journal={ISPRS Journal of Photogrammetry and Remote Sensing},
  volume={94},
  pages={70--79},
  year={2014}
}

@InProceedings{Qian2022,
  author  = {Qian, Guocheng and Li, Yuchen and Peng, Houwen and Mai, Jinjie and Hammoud, Hasan and Elhoseiny, Mohamed and Ghanem, Bernard},
  booktitle = {Advances in Neural Information Processing Systems},
  title   = {Pointnext: Revisiting pointnet++ with improved training and scaling strategies},
  year    = {2022},
  pages   = {23192--23204}
}

@InProceedings{Thomas2019,
  author    = {Thomas, Hugues and Qi, Charles R and Deschaud, Jean-Emmanuel and Marcotegui, Beatriz and Goulette, Fran{\c{c}}ois and Guibas, Leonidas J},
  booktitle = {Proceedings of the IEEE/CVF International Conference on Computer Vision},
  title     = {Kpconv: Flexible and deformable convolution for point clouds},
  year      = {2019},
  pages     = {6411--6420}
}

@InProceedings{Wang2019,
  author    = {Wang, Lei and Huang, Yuchun and Hou, Yaolin and Zhang, Shenman and Shan, Jie},
  booktitle = {Proceedings of the IEEE/CVF Conference on Computer Vision and Pattern Recognition},
  title     = {Graph attention convolution for point cloud semantic segmentation},
  year      = {2019},
  pages     = {10296--10305}
}

@Article{Wang2019a,
  author    = {Wang, Yue and Sun, Yongbin and Liu, Ziwei and Sarma, Sanjay E and Bronstein, Michael M and Solomon, Justin M},
  journal   = {ACM Transactions on Graphics},
  title     = {Dynamic graph cnn for learning on point clouds},
  year      = {2019},
  number    = {5},
  pages     = {1--12},
  volume    = {38}
}

@InProceedings{Wu2024,
  author    = {Wu, Xiaoyang and Jiang, Li and Wang, Peng-Shuai and Liu, Zhijian and Liu, Xihui and Qiao, Yu and Ouyang, Wanli and He, Tong and Zhao, Hengshuang},
  booktitle = {Proceedings of the IEEE/CVF Conference on Computer Vision and Pattern Recognition},
  title     = {Point Transformer V3: Simpler Faster Stronger},
  year      = {2024},
  pages     = {4840--4851}
}

@InProceedings{Xu2021,
  author    = {Xu, Mutian and Ding, Runyu and Zhao, Hengshuang and Qi, Xiaojuan},
  booktitle = {Proceedings of the IEEE/CVF Conference on Computer Vision and Pattern Recognition},
  title     = {Paconv: Position adaptive convolution with dynamic kernel assembling on point clouds},
  year      = {2021},
  pages     = {3173--3182}
}

@InProceedings{Zhang2023,
  author    = {Zhang, Nan and Pan, Zhiyi and Li, Thomas H and Gao, Wei and Li, Ge},
  booktitle = {Proceedings of the IEEE/CVF Conference on Computer Vision and Pattern Recognition},
  title     = {Improving graph representation for point cloud segmentation via attentive filtering},
  year      = {2023},
  pages     = {1244--1254},
}

@inproceedings{yu2022point,
  title={Point-bert: Pre-training 3d point cloud transformers with masked point modeling},
  author={Yu, Xumin and Tang, Lulu and Rao, Yongming and Huang, Tiejun and Zhou, Jie and Lu, Jiwen},
  booktitle={Proceedings of the IEEE/CVF Conference on Computer Vision and Pattern Recognition},
  pages={19313--19322},
  year={2022}
}

@inproceedings{yan2020pointasnl,
  title={Pointasnl: Robust point clouds processing using nonlocal neural networks with adaptive sampling},
  author={Yan, Xu and Zheng, Chaoda and Li, Zhen and Wang, Sheng and Cui, Shuguang},
  booktitle={Proceedings of the IEEE/CVF Conference on Computer Vision and Pattern Recognition},
  pages={5589--5598},
  year={2020}
}

@inproceedings{chen2023pointgpt,
  title={Pointgpt: Auto-regressively generative pre-training from point clouds},
  author={Chen, Guangyan and Wang, Meiling and Yang, Yi and Yu, Kai and Yuan, Li and Yue, Yufeng},
  booktitle={Advances in Neural Information Processing Systems},
  pages={29667--29679},
  year={2023}
}

@ARTICLE{Liang2023,
  author={Liang, Jiye and Du, Zijin and Liang, Jianqing and Yao, Kaixuan and Cao, Feilong},
  journal={IEEE Transactions on Pattern Analysis and Machine Intelligence}, 
  title={Long and Short-Range Dependency Graph Structure Learning Framework on Point Cloud}, 
  year={2023},
  volume={45},
  number={12},
  pages={14975-14989}
}

@ARTICLE{wen2024pointwavelet,
  author={Wen, Cheng and Long, Jianzhi and Yu, Baosheng and Tao, Dacheng},
  journal={IEEE Transactions on Neural Networks and Learning Systems}, 
  title={PointWavelet: Learning in Spectral Domain for 3-D Point Cloud Analysis}, 
  year={2025},
  volume={36},
  number={3},
  pages={4400-4412}
}

@InProceedings{Qi2017a,
  author  = {Qi, Charles Ruizhongtai and Yi, Li and Su, Hao and Guibas, Leonidas J},
  booktitle = {Advances in Neural Information Processing Systems},
  title   = {Pointnet++: Deep hierarchical feature learning on point sets in a metric space},
  pages={5105--5114},
  year    = {2017}
}

@InProceedings{Zhao2021,
  author    = {Zhao, Hengshuang and Jiang, Li and Jia, Jiaya and Torr, Philip HS and Koltun, Vladlen},
  booktitle = {Proceedings of the IEEE/CVF International Conference on Computer Vision},
  title     = {Point transformer},
  year      = {2021},
  pages     = {16259--16268}
}

@ARTICLE{Du2024,
  author={Du, Zijin and Liang, Jianqing and Liang, Jiye and Yao, Kaixuan and Cao, Feilong},
  journal={IEEE Transactions on Pattern Analysis and Machine Intelligence}, 
  title={Graph Regulation Network for Point Cloud Segmentation}, 
  year={2024},
  volume={46},
  number={12},
  pages={7940-7955}
}

@Article{Sohail2024,
  author    = {Sohail, Shahab Saquib and Himeur, Yassine and Kheddar, Hamza and Amira, Abbes and Fadli, Fodil and Atalla, Shadi and Copiaco, Abigail and Mansoor, Wathiq},
  journal   = {Information Fusion},
  title     = {Advancing 3D point cloud understanding through deep transfer learning: A comprehensive survey},
  year      = {2024},
  pages     = {102601}
}

@inproceedings{pang2022masked,
  title={Masked Autoencoders for Point Cloud Self-supervised Learning},
  author={Pang, Yatian and Wang, Wenxiao and Tay, Francis EH and Liu, Wei and Tian, Yonghong and Yuan, Li},
  booktitle={European Conference on Computer Vision},
  pages={604--621},
  year={2022}
}

@inproceedings{armeni20163d,
  title={3d semantic parsing of large-scale indoor spaces},
  author={Armeni, Iro and Sener, Ozan and Zamir, Amir R and Jiang, Helen and Brilakis, Ioannis and Fischer, Martin and Savarese, Silvio},
  booktitle={Proceedings of the IEEE/CVF Conference on Computer Vision and Pattern Recognition},
  pages={1534--1543},
  year={2016}
}

@inproceedings{uy2019revisiting,
  title={Revisiting point cloud classification: A new benchmark dataset and classification model on real-world data},
  author={Uy, Mikaela Angelina and Pham, Quang-Hieu and Hua, Binh-Son and Nguyen, Thanh and Yeung, Sai-Kit},
  booktitle={Proceedings of the IEEE/CVF International Conference on Computer Vision},
  pages={1588--1597},
  year={2019}
}

@inproceedings{wu20153d,
  title={3d shapenets: A deep representation for volumetric shapes},
  author={Wu, Zhirong and Song, Shuran and Khosla, Aditya and Yu, Fisher and Zhang, Linguang and Tang, Xiaoou and Xiao, Jianxiong},
  booktitle={Proceedings of the IEEE/CVF Conference on Computer Vision and Pattern Recognition},
  pages={1912--1920},
  year={2015}
}

@article{chang2015shapenet,
  title={Shapenet: An information-rich 3d model repository},
  author={Chang, Angel X and Funkhouser, Thomas and Guibas, Leonidas and Hanrahan, Pat and Huang, Qixing and Li, Zimo and Savarese, Silvio and Savva, Manolis and Song, Shuran and Su, Hao and others},
  journal={arXiv preprint arXiv:1512.03012},
  year={2015}
}

@inproceedings{wang2024fly,
  title={On-the-fly Point Feature Representation for Point Clouds Analysis},
  author={Wang, Jiangyi and Cheng, Zhongyao and Zhao, Na and Cheng, Jun and Yang, Xulei},
  booktitle={Proceedings of the 32nd ACM International Conference on Multimedia},
  pages={9204--9213},
  year={2024}
}

@inproceedings{he2025graph,
  title={Graph embedded contrastive learning for multi-view clustering},
  author={He, Hongqing and Xu, Jie and Wen, Guoqiu and Ren, Yazhou and Zhao, Na and Zhu, Xiaofeng},
  booktitle={Proceedings of the International Joint Conference on Artificial Intelligence},
  pages={5336--5344},
  year={2025}
}

@inproceedings{mo2022simple,
  title={Simple unsupervised graph representation learning},
  author={Mo, Yujie and Peng, Liang and Xu, Jie and Shi, Xiaoshuang and Zhu, Xiaofeng},
  booktitle={Proceedings of the AAAI Conference on Artificial Intelligence},
  pages={7797--7805},
  year={2022}
}

@inproceedings{sheng2022rethinking,
  title={Rethinking IoU-based optimization for single-stage 3D object detection},
  author={Sheng, Hualian and Cai, Sijia and Zhao, Na and Deng, Bing and Huang, Jianqiang and Hua, Xian-Sheng and Zhao, Min-Jian and Lee, Gim Hee},
  booktitle={European Conference on Computer Vision},
  pages={544--561},
  year={2022}
}

@article{sheng2025ct3d++,
  title={CT3D++: Improving 3D Object Detection with Keypoint-Induced Channel-wise Transformer},
  author={Sheng, Hualian and Cai, Sijia and Zhao, Na and Deng, Bing and Liang, Qiao and Zhao, Min-Jian and Ye, Jieping},
  journal={International Journal of Computer Vision},
  volume={133},
  number={7},
  pages={4817--4836},
  year={2025}
}

@inproceedings{xu2024investigating,
  title={Investigating and mitigating the side effects of noisy views for self-supervised clustering algorithms in practical multi-view scenarios},
  author={Xu, Jie and Ren, Yazhou and Wang, Xiaolong and Feng, Lei and Zhang, Zheng and Niu, Gang and Zhu, Xiaofeng},
  booktitle={Proceedings of the IEEE/CVF Conference on Computer Vision and Pattern Recognition},
  pages={22957--22966},
  year={2024}
}

@article{peng2023grlc,
  title={GRLC: Graph representation learning with constraints},
  author={Peng, Liang and Mo, Yujie and Xu, Jie and Shen, Jialie and Shi, Xiaoshuang and Li, Xiaoxiao and Shen, Heng Tao and Zhu, Xiaofeng},
  journal={IEEE Transactions on Neural Networks and Learning Systems},
  volume={35},
  number={6},
  pages={8609--8622},
  year={2023}
}

@ARTICLE{10833915,
  author={Chen, Jianpeng and Ling, Yawen and Xu, Jie and Ren, Yazhou and Huang, Shudong and Pu, Xiaorong and Hao, Zhifeng and Yu, Philip S. and He, Lifang},
  journal={IEEE Transactions on Neural Networks and Learning Systems}, 
  title={Variational Graph Generator for Multiview Graph Clustering}, 
  year={2025},
  volume={36},
  number={6},
  pages={11078-11091}
}

@inproceedings{panimages,
  title={How Do Images Align and Complement LiDAR? Towards a Harmonized Multi-modal 3D Panoptic Segmentation},
  author={Pan, Yining and Cui, Qiongjie and Yang, Xulei and Zhao, Na},
  booktitle={International Conference on Machine Learning},
  year={2025}
}

@inproceedings{wang2025uncertainty,
  title={Uncertainty Meets Diversity: A Comprehensive Active Learning Framework for Indoor 3D Object Detection},
  author={Wang, Jiangyi and Zhao, Na},
  booktitle={Proceedings of the IEEE/CVF Conference on Computer Vision and Pattern Recognition},
  pages={20329--20339},
  year={2025}
}

@ARTICLE{zhao2024sdcot++,
  author={Zhao, Na and Qian, Peisheng and Wu, Fang and Xu, Xun and Yang, Xulei and Lee, Gim Hee},
  journal={IEEE Transactions on Image Processing}, 
  title={SDCoT++: Improved Static-Dynamic Co-Teaching for Class-Incremental 3D Object Detection}, 
  year={2025},
  volume={34},
  number={},
  pages={4188-4202}
}

@inproceedings{xiu2025geometric,
  title={Geometric Alignment and Prior Modulation for View-Guided Point Cloud Completion on Unseen Categories},
  author={Xiu, Jingqiao and Li, Yicong and Zhao, Na and Fang, Han and Wang, Xiang and Yao, Angela},
  booktitle={Proceedings of the IEEE/CVF International Conference on Computer Vision},
  pages={27435--27444},
  year={2025}
}

@inproceedings{jianggaussianblock,
  title={GaussianBlock: Building Part-Aware Compositional and Editable 3D Scene by Primitives and Gaussians},
  author={Jiang, Shuyi and Zhao, Qihao and Rahmani, Hossein and Soh, De Wen and Liu, Jun and Zhao, Na},
  booktitle={International Conference on Learning Representations},
  year={2025}
}

@inproceedings{li2024end,
  title={End-to-end semi-supervised 3d instance segmentation with pcteacher},
  author={Li, Linfeng and Zhao, Na},
  booktitle={IEEE International Conference on Robotics and Automation},
  pages={5352--5358},
  year={2024}
}

@inproceedings{han2024dual,
  title={Dual-perspective knowledge enrichment for semi-supervised 3d object detection},
  author={Han, Yucheng and Zhao, Na and Chen, Weiling and Ma, Keng Teck and Zhang, Hanwang},
  booktitle={Proceedings of the AAAI Conference on Artificial Intelligence},
  pages={2049--2057},
  year={2024}
}

@inproceedings{li2024laso,
  title={Laso: Language-guided affordance segmentation on 3d object},
  author={Li, Yicong and Zhao, Na and Xiao, Junbin and Feng, Chun and Wang, Xiang and Chua, Tat-seng},
  booktitle={Proceedings of the IEEE/CVF Conference on Computer Vision and Pattern Recognition},
  pages={14251--14260},
  year={2024}
}

@inproceedings{wang2025augrefer,
  title={Augrefer: Advancing 3d visual grounding via cross-modal augmentation and spatial relation-based referring},
  author={Wang, Xinyi and Zhao, Na and Han, Zhiyuan and Guo, Dan and Yang, Xun},
  booktitle={Proceedings of the AAAI Conference on Artificial Intelligence},
  pages={8006--8014},
  year={2025}
}

@inproceedings{wangaffordbot,
  title={AffordBot: 3D Fine-grained Embodied Reasoning via Multimodal Large Language Models},
  author={Wang, Xinyi and Yang, Xun and Xu, Yanlong and Wu, Yuchen and Li, Zhen and Zhao, Na},
  booktitle={Advances in Neural Information Processing Systems},
  year={2025}
}

@article{yuan2025scene,
  title={Scene-R1: Video-Grounded Large Language Models for 3D Scene Reasoning without 3D Annotations},
  author={Yuan, Zhihao and Jiang, Shuyi and Feng, Chun-Mei and Zhang, Yaolun and Cui, Shuguang and Li, Zhen and Zhao, Na},
  journal={arXiv preprint arXiv:2506.17545},
  year={2025}
}

@inproceedings{wu2023sketch,
  title={Sketch and text guided diffusion model for colored point cloud generation},
  author={Wu, Zijie and Wang, Yaonan and Feng, Mingtao and Xie, He and Mian, Ajmal},
  booktitle={Proceedings of the IEEE/CVF International Conference on Computer Vision},
  pages={8929--8939},
  year={2023}
}

@String(AAAI = {AAAI})

\end{document}